\documentclass{article}

\usepackage[preprint]{neurips_2026}

\usepackage[utf8]{inputenc} 
\usepackage[T1]{fontenc}    
\usepackage{hyperref}       
\usepackage{url}            
\usepackage{booktabs}       
\usepackage{amsfonts}       
\usepackage{xcolor}         
\usepackage{graphicx}       
\usepackage{amsmath}        

\title{Can LLMs Predict Polymer Physics Just by Reading Synthesis and Processing Prose?}

\author{%
  Yuchu Liu\thanks{Equal contribution.} \\
  Yale University \\
  \And
  Rui Zhu\footnotemark[1] \\
  Yale University \\
  \And
  Jingwei Xiong \\
  UC Davis \\
  \And
  Haixu Tang\thanks{Corresponding author.} \\
  Indiana University Bloomington \\
}

\begin{document}

\maketitle

\begin{abstract}
Can large language models predict physical and mechanical polymer properties simply by reading unstructured scientific prose? Polymer performance is rarely determined by chemical structure alone; identical nominal polymers can exhibit drastically different behaviors depending on their synthesis route, processing history, morphology, and testing conditions. Yet, state-of-the-art polymer property models typically rely on structure-only representations---such as SMILES or molecular graphs---which strip away this vital experimental context. In this work, we introduce \textbf{PolyLM}, a natural-language-only, process- and condition-aware framework that predicts materials performance directly from full-text literature. By circumventing structural inputs entirely, PolyLM preserves the nuanced, unstructured descriptions of synthesis and processing reported by domain scientists. To train this framework, we curated an unprecedented, literature-scale dataset encompassing 185,000 scientific papers and over 276,400 unique polymer samples across 22 physical, mechanical, and thermal properties. We fine-tuned a massive 9-billion-parameter language model (Qwen3.5-9B) using Low-Rank Adaptation (LoRA) and task-level uncertainty weighting. Evaluated on 68,283 held-out observations, the model achieves remarkably high predictive accuracy, establishing new state-of-the-art benchmarks for complex properties. Across the 22 diverse targets, the model achieves a median $R^2$ of 0.74, with predictions for key thermal, mechanical, and physicochemical properties frequently surpassing an $R^2$ of 0.80. Furthermore, a rigorous matched ablation study confirms that preserving natural-language processing context is essential, preventing an average $R^2$ collapse of 0.062 across mechanical properties. Moreover, we demonstrate that state-of-the-art generalist frontier models (such as Claude 3 Opus 4.6) fail completely at zero-shot physical regression, underscoring the necessity of our specialized fine-tuning approach. These results unequivocally demonstrate that natural language is a powerful, highly scalable interface for realistic materials performance prediction.
\end{abstract}

\section{Introduction}

\begin{figure}[htbp]
    \centering
    \includegraphics[width=\textwidth]{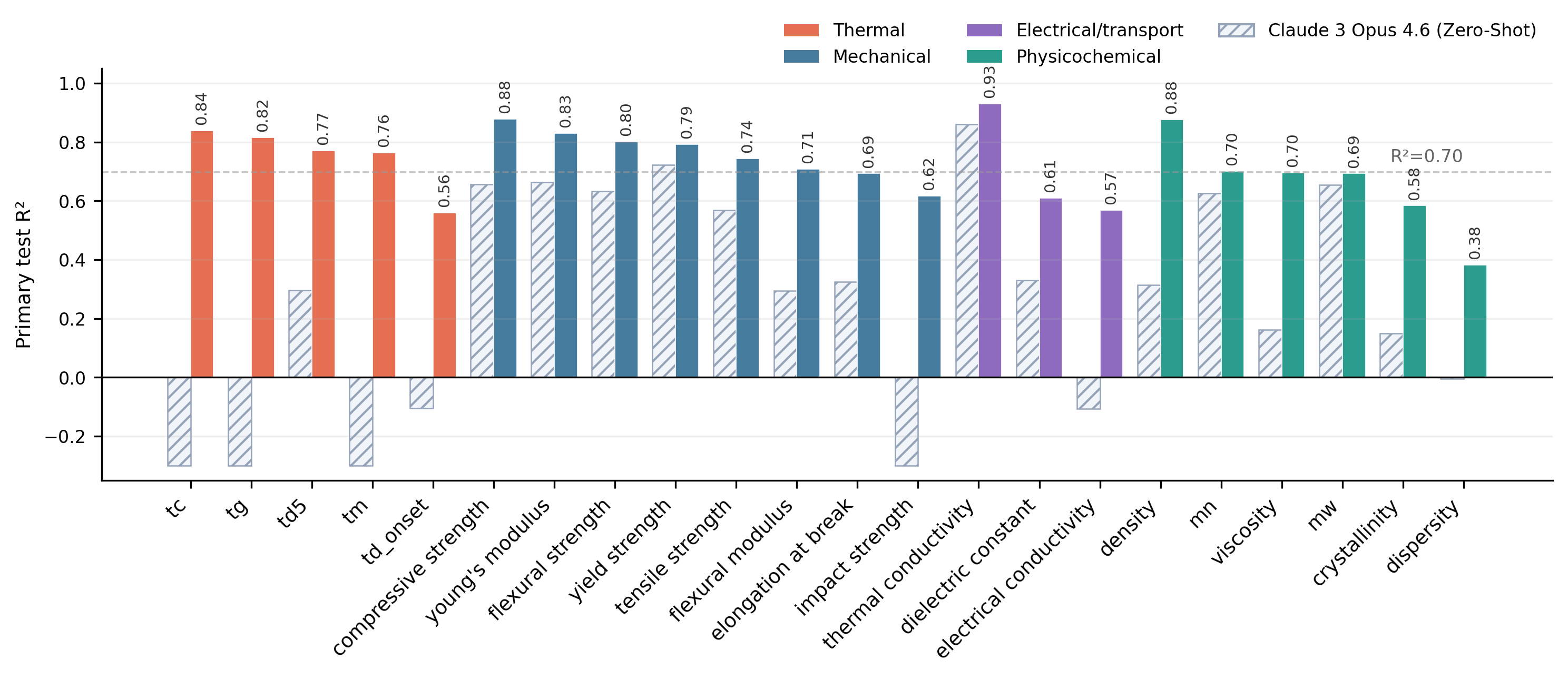}
    \caption{Primary predictive performance (measured by held-out test $R^2$) of the unified PolyLM model across 22 complex properties. The shaded bars indicate the zero-shot performance of Claude 3 Opus 4.6 (clipped at 0 or extended to -0.3 for negative $R^2$ values) on the identical strict numeric-only subset, demonstrating PolyLM's substantial superiority over unspecialized frontier models. For targets spanning multiple orders of magnitude (e.g., modulus, conductivity), properties were regressed in logarithmic space, and the primary $R^2$ represents the log-space fit. See Table~\ref{tab:unified_results} for complete regression metrics.}
    \label{fig:unified_r2_teaser}
\end{figure}

Polymer property prediction is a cornerstone for accelerating materials discovery. However, real-world polymer performance is not simply a deterministic function of a chemical repeat unit. The measurable attributes of a material emerge from a complex interplay of chemistry, composition, processing, morphology, and testing protocols. Fundamental properties such as glass transition temperature ($T_g$), melting temperature ($T_m$), degradation behavior, modulus, strength, elongation, conductivity, density, and viscosity can shift dramatically based on molecular weight distribution, additives, curing routes, thermal history, strain rates, and measurement temperatures. In the primary scientific literature, these crucial contextual factors are typically reported in unstructured prose rather than standardized machine-readable formats.

This reality creates a fundamental mismatch between many existing polymer-property models and the data abundantly available in full-text scientific reports. Current state-of-the-art structure-based representations---such as SMILES strings \cite{weininger1988smiles}, molecular fingerprints, graph neural networks (GNNs) \cite{gilmer2017neural}, PSELFIES \cite{savit2025polybart}, and serialized polymer graphs---provide exceptionally strong baselines when clean, unambiguous structures are available. Yet, many experimentally synthesized polymers, complex blends, composites, heavily crosslinked networks, and post-processed specimens lack a unique or readily discernible SMILES representation. More importantly, even when a correct repeat-unit representation can be assigned, it inherently omits the processing and condition variables that dictate the ultimately measured performance. For many physical and mechanical properties, this missing context is not nuisance metadata but the primary predictive signal.

To address this representation gap, we introduce \textbf{PolyLM}, a framework that treats natural language as the primary modeling interface. Rather than reducing each material to a rigid molecular string, PolyLM preserves the nuanced textual context that accompanies experimental measurements: the detailed sample description, composition, synthesis and processing routes, and specific measurement conditions. The framework is inherently \textit{natural-language-only} (eliminating the strict requirement for SMILES) and \textit{process- and condition-aware} (explicitly conditioning predictions on how a material was fabricated, processed, and tested). We augment existing benchmark databases \cite{ma2020pi1m, otsuka2011polyinfo} by synthesizing vast amounts of unstructured literature data.

Our central claim is that a literature-scale, context-rich natural-language formulation enables realistic materials performance prediction that better reflects experimental realities. We substantiate this claim through the development of a unified 22-property model, rigorous controlled ablations of process-aware inputs, extensive extraction-quality audits, and a thorough task-level uncertainty analysis.

Our key contributions are as follows:
\begin{enumerate}
    \item \textbf{A Literature-Scale Natural-Language Polymer Regression Corpus:} We developed an automated full-text extraction pipeline over 524 literature shards (roughly 185,000 papers), aggregating approximately 276,400 unique polymer samples across 22 distinct target properties.
    \item \textbf{A Unified 22-Property PolyLM Regressor:} We engineered a 9B-parameter language-model backbone augmented with LoRA regression adapters and property-specific heads, capable of predicting thermal, mechanical, electrical/transport, and physicochemical properties directly from context-rich natural-language inputs.
    \item \textbf{A Controlled Test of Process-Aware Input:} Through a strictly matched mechanical property ablation, we quantitatively demonstrate that removing synthesis and processing context uniformly degrades performance across all 8 mechanical properties, reducing the mean primary $R^2$ by 0.062.
    \item \textbf{A Task-Level Uncertainty Benchmark:} We analyze learned per-property uncertainty weights, showing they partially rank task difficulty, while identifying limitations regarding their use as calibrated sample-level intervals.
    \item \textbf{A Zero-Shot Frontier Model Benchmark:} We evaluate state-of-the-art conversational LLMs (Claude 3 Opus 4.6 and Qwen3.5-9B) on continuous property prediction. The resulting negative $R^2$ values demonstrate that massive unspecialized language models cannot reliably perform condition-aware physical regression without our proposed task-specific fine-tuning framework.
\end{enumerate}

\begin{figure}[htbp]
    \centering
    \includegraphics[width=\linewidth]{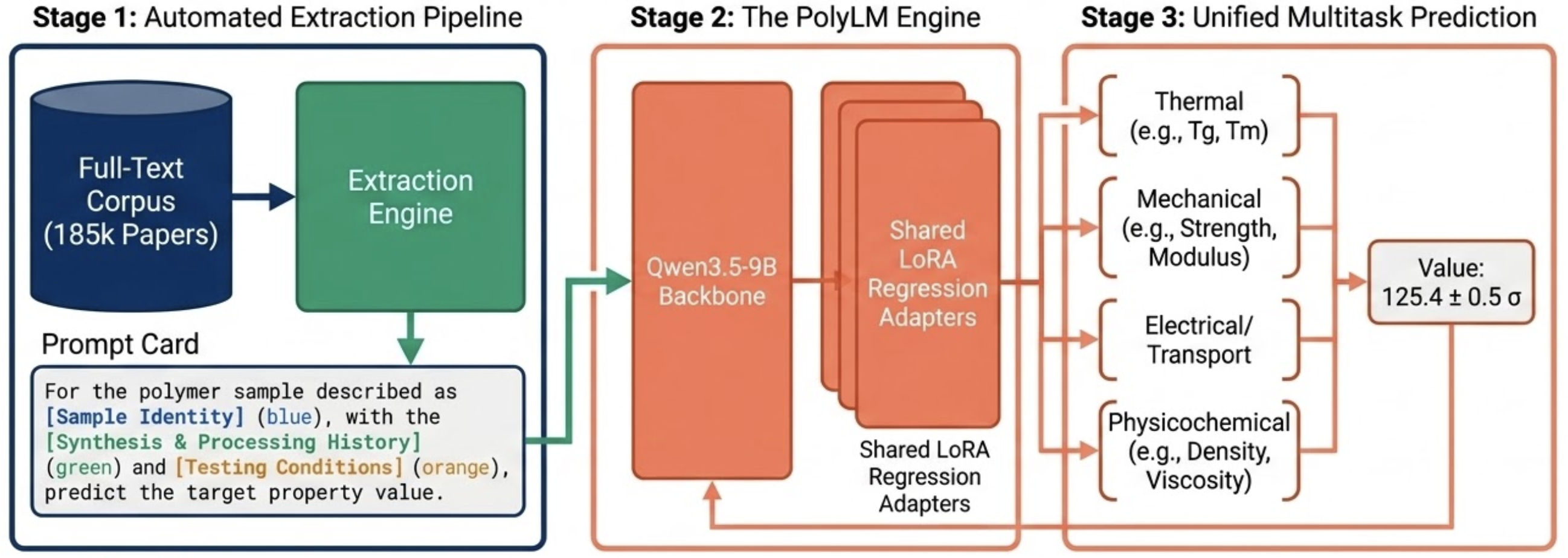}
    \caption{The PolyLM architecture and processing pipeline. The framework automatically extracts context-rich natural-language prompts from $\sim$185k scientific papers, explicitly preserving vital synthesis and processing history without relying on molecular graphs (Stage 1). These textual inputs are processed by a Qwen3.5-9B backbone equipped with shared LoRA regression adapters (Stage 2) to perform unified multitask prediction. The model routes representations to distinct property heads, outputting continuous values and learned uncertainty estimates ($\sigma$) across 22 diverse polymer targets (Stage 3).}
    \label{fig:polylm_pipeline}
\end{figure}

\section{Related Work}

\subsection{Structure-Based Polymer-Property Prediction}
A substantial body of research in polymer informatics \cite{pilania2021machine, chen2021polymer} has focused on representing materials using molecular fingerprints, graph neural networks, SMILES-derived embeddings, and more recently, specialized string formats like SELFIES \cite{krenn2020selfies}, PSELFIES \cite{savit2025polybart}, or serialized polymer graphs. Transformer-based models like ChemBERTa \cite{chithrananda2020chemberta} have also adapted structural inputs for large-scale pretraining. These approaches establish strong baselines and excel in curated environments where idealized repeat-unit structures are readily available. However, their efficacy diminishes when applied to literature-scale databases. Real-world samples encompass complex compositions, intricate processing histories, and diverse measurement environments that are difficult, if not impossible, to encode comprehensively within a purely structural molecular representation. 

\subsection{Language Models for Materials Literature}
The application of natural language processing to materials science has seen rapid growth, primarily centered on named-entity recognition (NER) \cite{weston2019named}, relation extraction, information retrieval, and scientific question answering \cite{olivetti2020data}. Unsupervised language models have successfully captured latent knowledge representations \cite{tshitoyan2019unsupervised}. Domain-specific models like MaterialsBERT \cite{gupta2022matscibert} have demonstrated that language models can effectively parse and comprehend materials text. PolyLM diverges from these extraction-focused paradigms by utilizing the full-text extracted natural-language descriptions not merely as an intermediate step for building structured databases, but directly as the input representation for downstream continuous property regression. This end-to-end approach leverages the intrinsic capacity of large language models to contextualize unstructured scientific prose \cite{li2025knowledge, li2023stinmatch}.

\subsection{Context-Aware Materials Performance Prediction}
Processing-structure-property relationships are the fundamental paradigm of materials science \cite{ramprasad2017machine}, yet a disproportionate share of machine-learning benchmarks focus strictly on structure-property mappings. For polymers, processing variables critically dictate latent variables like crystallinity, morphology, and crosslink density that exert immense influence on macroscopic performance. PolyLM is architected around this physical reality. By retaining synthesis, processing, and conditional information as active predictive inputs rather than discarding them as noisy metadata during curation, PolyLM bridges the gap between idealized molecular models and experimental materials science.

\section{Data and Extraction Pipeline}

\subsection{Full-Text Source and Extraction}
The PolyLM corpus is synthesized entirely from full-text scientific literature. We developed a robust extraction pipeline that processes 524 shards, corresponding to approximately 185,000 polymer-relevant publications. For each paper, the pipeline automatically identifies distinct material samples, extracts comprehensive sample descriptions along with synthesis and processing procedures, and robustly maps reported experimental measurements to normalized property heads and standard units.

Each extracted training instance is formulated as a natural-language prompt. At a minimum, this input comprises a material or sample description. When available, it is augmented with synthesis routes, processing conditions, compositional details, treatments, and measurement parameters. This format purposefully preserves experimentally relevant context that is typically discarded in structure-only featurization schemes.

\subsection{Target Properties}
Our unified framework models 22 distinct properties categorized into four broad domains:
\begin{itemize}
    \item \textbf{Thermal:} Glass transition temperature ($T_g$), melting temperature ($T_m$), 5\% weight loss temperature ($T_{d5}$), degradation onset temperature.
    \item \textbf{Mechanical:} Tensile strength, Young's modulus, elongation at break, flexural strength, compressive strength, impact strength, yield strength, flexural modulus.
    \item \textbf{Electrical/Transport:} Electrical conductivity, dielectric constant, thermal conductivity.
    \item \textbf{Physicochemical:} Density, number-average molecular weight ($M_n$), weight-average molecular weight ($M_w$), dispersity, crystallinity, viscosity.
\end{itemize}
The rigorously held-out test set comprises 68,283 property observations across these 22 heads. Properties with value distributions spanning several orders of magnitude (e.g., modulus, viscosity, conductivity) are trained and evaluated in log space to stabilize gradients and provide more meaningful performance metrics; for these targets, log-space $R^2$ is our primary evaluation metric.

\subsection{Extraction Quality Audit}
To rigorously quantify the fidelity of our extraction pipeline, we conducted a manual annotation audit of 120 parseable property records. The results, summarized in Table~\ref{tab:extraction_audit}, demonstrate high precision across critical dimensions. The audit yielded a sample-association precision of 1.000, property-mapping precision of 0.908, value precision of 0.942, and unit precision of 0.942. The strict record precision (requiring all components to be flawlessly extracted) achieved 0.842. Identified error modes primarily consisted of underspecified property labels, malformed limits, missed numeric ranges, omitted units, and complex equation-like relationships erroneously reduced to single coefficients.

\begin{table}[htbp]
    \caption{Extraction Quality Audit. Manual annotation of 120 randomly sampled property records demonstrates high fidelity in extracting values, units, and sample associations from unstructured full-text literature.}
    \label{tab:extraction_audit}
    \centering
    \begin{tabular}{lcc}
        \toprule
        \textbf{Extraction Component} & \textbf{Correct / Total} & \textbf{Precision} \\
        \midrule
        Sample Association & 120 / 120 & 1.000 \\
        Property Mapping & 109 / 120 & 0.908 \\
        Value Extraction & 113 / 120 & 0.942 \\
        Unit Extraction & 113 / 120 & 0.942 \\
        \midrule
        \textbf{Strict Record Precision} & \textbf{101 / 120} & \textbf{0.842} \\
        \bottomrule
    \end{tabular}
\end{table}

\section{Model}

The defining architectural choice of PolyLM is its exclusive reliance on natural language as the continuous representation interface, circumventing the need for rigid molecular graphs or strings. As illustrated in Figure~\ref{fig:polylm_pipeline}, the framework operates in three interconnected stages: context-rich prompt formulation, large language model (LLM) representation via parameter-efficient tuning, and multi-task routing through property-specific regression heads.

\subsection{Stage 1: Context-Rich Prompt Formulation}
To ensure that physical measurements are accurately tied to their specific experimental conditions, PolyLM ingests unstructured text directly. Each training instance is constructed using a strict templating protocol comprising a \texttt{[Sample]} block and a \texttt{[Synthesis]} block. The \texttt{[Sample]} block delineates the material's identity, including copolymers, blends, and exact compositing ratios. The \texttt{[Synthesis]} block captures the fabrication history, such as curing times, extrusion temperatures, molecular weight metrics, and physical testing conditions. Crucially, the target properties are aggressively masked during this stage to prevent label leakage. This formulation allows the self-attention mechanisms of the LLM to dynamically condition predictions on the exact processing nuances that dictate measurable performance, a capability fundamentally absent in structure-only representations.

\subsection{Stage 2: LLM Representation and Parameter-Efficient Tuning}
The core representation engine of PolyLM is built upon a 9B-parameter causal language model backbone \cite{vaswani2017attention, devlin2018bert, brown2020language, raffel2020exploring, touvron2023llama}. Specifically, the production model leverages Qwen3.5-9B \cite{bai2023qwen}, which was initialized from an intermediate checkpoint following extensive domain-adaptive pretraining on a massive corpus of polymer literature.

To facilitate unified, multi-property continuous regression without catastrophically forgetting its underlying chemical reasoning capabilities, we employ Low-Rank Adaptation (LoRA) \cite{hu2021lora}. We inject trainable low-rank matrices into all primary self-attention and feed-forward projection layers (i.e., \texttt{q\_proj}, \texttt{k\_proj}, \texttt{v\_proj}, \texttt{up\_proj}, \texttt{down\_proj}, and \texttt{gate\_proj}). Furthermore, we utilize a strategic layer-freezing approach: the bottom transformer blocks remain entirely frozen to preserve foundational semantic extraction, while the uppermost layers are fully tunable. This approach drastically reduces the trainable parameter count to less than 2\% of the total model size, enabling efficient training on literature-scale datasets while maintaining high representational capacity.

\subsection{Stage 3: Multi-Task Routing and Regression Heads}
Translating a sequence of contextual text tokens into 22 distinct continuous physical properties requires a robust pooling and routing mechanism. First, the final hidden states of the LLM are aggregated into a single fixed-length global vector. While standard mean-pooling over non-padding tokens is effective, PolyLM also utilizes a learnable attention-pooling mechanism that computes dynamic token weights, allowing the model to focus strictly on physical units and quantitative conditions rather than filler syntax.

This pooled representation is subsequently routed through a deep, shared multi-layer perceptron (MLP) trunk consisting of linear projections, Layer Normalization, GELU activations, and sequential Residual Blocks. This trunk creates a compressed, high-order representation of the material's physical state (bottlenecked to 128 dimensions). Finally, 22 independent, property-specific linear regression heads branch off from this shared trunk to output the final scalar predictions.

\subsection{Optimization and Density-Weighted Loss}
Training a unified model on highly heterogeneous polymer data introduces severe optimization challenges. The 22 target properties exhibit drastically different dynamic ranges and literature sparsities. Consequently, properties spanning multiple orders of magnitude (e.g., modulus, viscosity) are regressed in logarithmic space, and all labels undergo property-specific $z$-score normalization based on the training distribution. During gradient accumulation, missing properties for a given sample are masked out via NaN-filtering, ensuring that the unified model only updates weights for explicitly observed targets.

More importantly, physical property distributions in the literature are notoriously skewed; standard or mediocre materials dominate the corpus, while exceptional materials (the primary targets of discovery) occupy the long tails. To prevent the model from collapsing to the mean, PolyLM implements an inverse-density weighted Mean Squared Error (MSE) objective. Let $y_i$ be the normalized target value for sample $i$ and $\hat{y}_i$ be the model prediction. We apply Gaussian Kernel Density Estimation (KDE) over the training distribution to estimate the local probability density $p(y_i)$. The KDE-weighted loss $\mathcal{L}_{\text{KDE}}$ for a single property task $t$ over $N$ valid samples is defined as:

\begin{equation}
    \mathcal{L}_{\text{KDE}}^{(t)} = \frac{1}{N} \sum_{i=1}^{N} w(y_i) \left( \hat{y}_i^{(t)} - y_i^{(t)} \right)^2
\end{equation}

where the per-sample weight $w(y_i)$ is inversely proportional to the density, $w(y_i) \propto 1 / \max(p(y_i), \epsilon)$, truncated by a threshold $\epsilon$ (the 5th percentile of the density distribution) to prevent exploding gradients for extreme outliers, and normalized such that $\mathbb{E}[w(y_i)] = 1$.

Finally, to unify the 22 tasks, this density-weighted objective is augmented by a multi-task homoscedastic uncertainty weighting \cite{kendall2018multi}. We introduce a learned variance parameter $\sigma_t^2$ for each property $t \in \{1, \dots, 22\}$. The total optimization objective $\mathcal{L}_{\text{total}}$ dynamically balances the gradients across all tasks:

\begin{equation}
    \mathcal{L}_{\text{total}} = \sum_{t=1}^{22} \left( \frac{1}{2\sigma_t^2} \mathcal{L}_{\text{KDE}}^{(t)} + \log(\sigma_t) \right)
\end{equation}

This formulation automatically suppresses noisier tasks during training and provides a valuable task-level uncertainty heuristic ($\sigma_t$), which correlates with the intrinsic difficulty of predicting each property from literature text (further analyzed in Section~\ref{sec:uncertainty}).

\section{Experiments and Results}

Our empirical evaluation is designed to answer the core hypothesis of this work: can a natural-language representation that faithfully preserves processing and condition context support robust, broad-scale polymer performance prediction? We first establish the efficacy of the unified 22-property model, then isolate the specific value of process and synthesis text through a controlled ablation, analyze the learned task-level uncertainty, and conclude with an external literature benchmark.

\subsection{Unified 22-Property Performance}

Table~\ref{tab:unified_results} and Figure~\ref{fig:unified_r2_teaser} summarize the performance of the v3 unified 22-property model. Evaluated on an expansive held-out test set comprising 68,283 property observations, PolyLM achieves a median primary $R^2$ of 0.744. Notably, 13 out of 22 properties exceed a primary $R^2$ of 0.70, and 7 surpass 0.80. This demonstrates that a single, unified natural-language model can concurrently master a highly diverse set of polymer performance targets.

The most predictive heads span multiple physical domains. For instance, the model accurately predicts common thermal properties such as $T_g$ ($R^2 = 0.826$, $n=5,391$) and density ($R^2 = 0.878$, $n=2,155$), as well as complex, context-sensitive mechanical properties like log-space compressive strength ($R^2_{\log} = 0.878$, $n=1,537$) and log-space Young's modulus ($R^2_{\log} = 0.830$, $n=9,777$). 

Performance naturally degrades for targets characterized by immense dynamic ranges, highly sparse literature labels, or inherently heterogeneous reporting conventions—such as electrical conductivity, dielectric constant, dispersity, and viscosity. These underperforming heads are scientifically instructive; they highlight the frontiers where literature curation, sophisticated condition extraction, and physical modeling remain significant bottlenecks.

\begin{table}[htbp]
    \caption{Unified 22-Property Held-Out Results. Primary metrics are indicated for each target, utilizing log-space $R^2$ for properties spanning multiple orders of magnitude.}
    \label{tab:unified_results}
    \centering
    \resizebox{\textwidth}{!}{
    \begin{tabular}{llrrrrr}
        \toprule
        \textbf{Group} & \textbf{Property} & \textbf{Test $n$} & \textbf{$R^2$ (linear)} & \textbf{$R^2$ (log)} & \textbf{MAE} & \textbf{RMSE} \\
        \midrule
        Thermal & $T_g$ & 5,391 & \textbf{0.826} & 0.816 & 18.76 & 31.92 \\
        Thermal & $T_m$ & 4,066 & \textbf{0.744} & 0.763 & 16.42 & 35.00 \\
        Thermal & $T_c$ & 1,437 & \textbf{0.832} & 0.839 & 12.59 & 22.99 \\
        Thermal & $T_{d5}$ & 1,685 & \textbf{0.809} & 0.770 & 26.02 & 40.84 \\
        Thermal & $T_d$ onset & 2,765 & \textbf{0.588} & 0.559 & 38.99 & 60.04 \\
        \midrule
        Mechanical & Tensile strength & 12,819 & 0.228 & \textbf{0.744} & 44.94 & 463.36 \\
        Mechanical & Young's modulus & 9,777 & 0.573 & \textbf{0.830} & 2428.21 & 13023.20 \\
        Mechanical & Elongation at break & 8,886 & \textbf{0.693} & --- & 72.99 & 145.99 \\
        Mechanical & Flexural strength & 2,862 & 0.497 & \textbf{0.801} & 41.12 & 186.97 \\
        Mechanical & Compressive strength & 1,537 & 0.621 & \textbf{0.878} & 23.43 & 98.46 \\
        Mechanical & Impact strength & 1,012 & \textbf{0.688} & 0.617 & 10.13 & 26.59 \\
        Mechanical & Yield strength & 782 & 0.060 & \textbf{0.793} & 27.39 & 263.99 \\
        Mechanical & Flexural modulus & 1,721 & 0.559 & \textbf{0.708} & 3242.65 & 10400.47 \\
        \midrule
        Electrical/Transport & Electrical conductivity & 1,054 & 0.002 & \textbf{0.569} & $9.15\times 10^4$ & $1.43\times 10^6$ \\
        Electrical/Transport & Dielectric constant & 734 & 0.100 & \textbf{0.610} & 49.60 & 412.15 \\
        Electrical/Transport & Thermal conductivity & 144 & 0.442 & \textbf{0.930} & 18.50 & 62.59 \\
        \midrule
        Physicochemical & Density & 2,155 & \textbf{0.878} & --- & 0.14 & 0.29 \\
        Physicochemical & $M_n$ & 1,359 & 0.145 & \textbf{0.701} & 60691.71 & 650576.16 \\
        Physicochemical & $M_w$ & 1,055 & 0.048 & \textbf{0.693} & 609606.23 & 5203490.02 \\
        Physicochemical & Dispersity & 2,025 & \textbf{0.381} & --- & 0.90 & 2.66 \\
        Physicochemical & Crystallinity & 3,736 & \textbf{0.584} & --- & 10.08 & 15.00 \\
        Physicochemical & Viscosity & 1,281 & 0.069 & \textbf{0.696} & 38232.26 & 836646.77 \\
        \bottomrule
    \end{tabular}
    }
\end{table}

\subsection{Zero-Shot Baselines: Qwen3.5-9B and Claude 3 Opus}
In addition to standard regression evaluation, we assessed whether the underlying language models could inherently perform continuous physical property prediction without our fine-tuning paradigm. We evaluated the base Qwen3.5-9B model in a zero-shot setting across all 22 tasks. The performance collapsed entirely, yielding a macro-averaged $R^2$ of $-1.0817$ and a Mean Absolute Error (MAE) of $84.4\times 10^3$ K (for temperature-based targets), indicating that the foundation model produces effectively random numerical artifacts out-of-the-box.

To verify whether this limitation extends to state-of-the-art frontier models, we additionally evaluated Anthropic's Claude 3 Opus 4.6 \cite{anthropic2024claude} in a strict zero-shot regime. Because conversational models frequently output ranges or explanatory text, we implemented a rigorous parsing pipeline to isolate strict numeric responses, which retained 83.9\% of the test instances. Even on this favorable subset, Claude 3 Opus 4.6 produced a macro $R^2$ of $-0.987$ and a macro log-$R^2$ of $-2.269$ on the unified suite, drastically underperforming PolyLM (macro $R^2 = 0.471$, log-$R^2 = 0.740$). While Opus occasionally demonstrated a weak positive linear signal on 5 out of 44 property configurations (such as dielectric constant and thermal conductivity), PolyLM definitively outperformed it on all comparable log-$R^2$ evaluations. This confirms that while massive frontier models possess some latent domain knowledge, they cannot reliably substitute for specialized, condition-aware regression architectures.

\subsection{Process and Synthesis Input Ablation}

To rigorously quantify the value of process-aware modeling, we conducted the EXP-A1 matched ablation study (Figure~\ref{fig:ablation} and Table~\ref{tab:ablation}). This experiment tests whether processing and synthesis descriptions provide actionable predictive signal beyond the baseline material description. We utilized a strictly controlled setting: identical language-model backbone (Checkpoint B), identical data splits, matching LoRA configurations, and fixed training hyperparameters. The sole variable was the input representation: \texttt{sample\_synthesis} (full natural language including synthesis and processing text) versus \texttt{sample\_only} (a scrubbed input utilizing only the sample field, with synthesis removed).

The results are definitive. Removing synthesis and processing information uniformly degrades predictive performance across all 8 mechanical properties. The mean primary $R^2$ drops significantly from 0.723 to 0.661 ($\Delta = -0.062$). The most severe degradations are observed in impact strength ($\Delta = -0.084$), yield strength ($\Delta = -0.080$), tensile strength ($\Delta = -0.078$), and elongation at break ($\Delta = -0.065$). Tellingly, not a single mechanical property head benefited from the \texttt{sample\_only} formulation.

This ablation provides the critical evidence supporting our central claim. Mechanical properties are notoriously sensitive to sample history, thermal treatments, morphology, filler dispersion, and testing conditions. The consistent performance collapse under the \texttt{sample\_only} condition mathematically confirms that synthesis and processing descriptions carry vital predictive signals and must not be discarded as auxiliary metadata.

\begin{table}[htbp]
    \caption{EXP-A1 Mechanical Input Ablation. Comparing full synthesis context against a sample-only baseline. Removing process history uniformly degrades mechanical property predictions.}
    \label{tab:ablation}
    \centering
    \resizebox{\columnwidth}{!}{
    \begin{tabular}{lrlrrr}
        \toprule
        \textbf{Property} & \textbf{Test $n$} & \textbf{Metric} & \textbf{Sample+Synthesis} & \textbf{Sample Only} & \textbf{$\Delta$} \\
        \midrule
        Tensile strength & 14,504 & $R^2$ (log) & 0.767 & 0.689 & -0.078 \\
        Young's modulus & 11,509 & $R^2$ (log) & 0.831 & 0.782 & -0.049 \\
        Elongation at break & 9,899 & $R^2$ (linear) & 0.638 & 0.573 & -0.065 \\
        Flexural strength & 3,595 & $R^2$ (log) & 0.773 & 0.741 & -0.032 \\
        Flexural modulus & 2,063 & $R^2$ (log) & 0.719 & 0.663 & -0.056 \\
        Compressive strength & 1,618 & $R^2$ (log) & 0.846 & 0.794 & -0.052 \\
        Impact strength & 1,188 & $R^2$ (log) & 0.501 & 0.417 & -0.084 \\
        Yield strength & 1,074 & $R^2$ (log) & 0.709 & 0.629 & -0.080 \\
        \midrule
        \textbf{Mean} & --- & \textbf{Primary} & \textbf{0.723} & \textbf{0.661} & \textbf{-0.062} \\
        \bottomrule
    \end{tabular}
    }
\end{table}

\subsection{Task-Level Uncertainty}
\label{sec:uncertainty}

\begin{figure}[htbp]
    \centering
    \begin{minipage}[t]{0.48\textwidth}
        \centering
        \includegraphics[width=\linewidth]{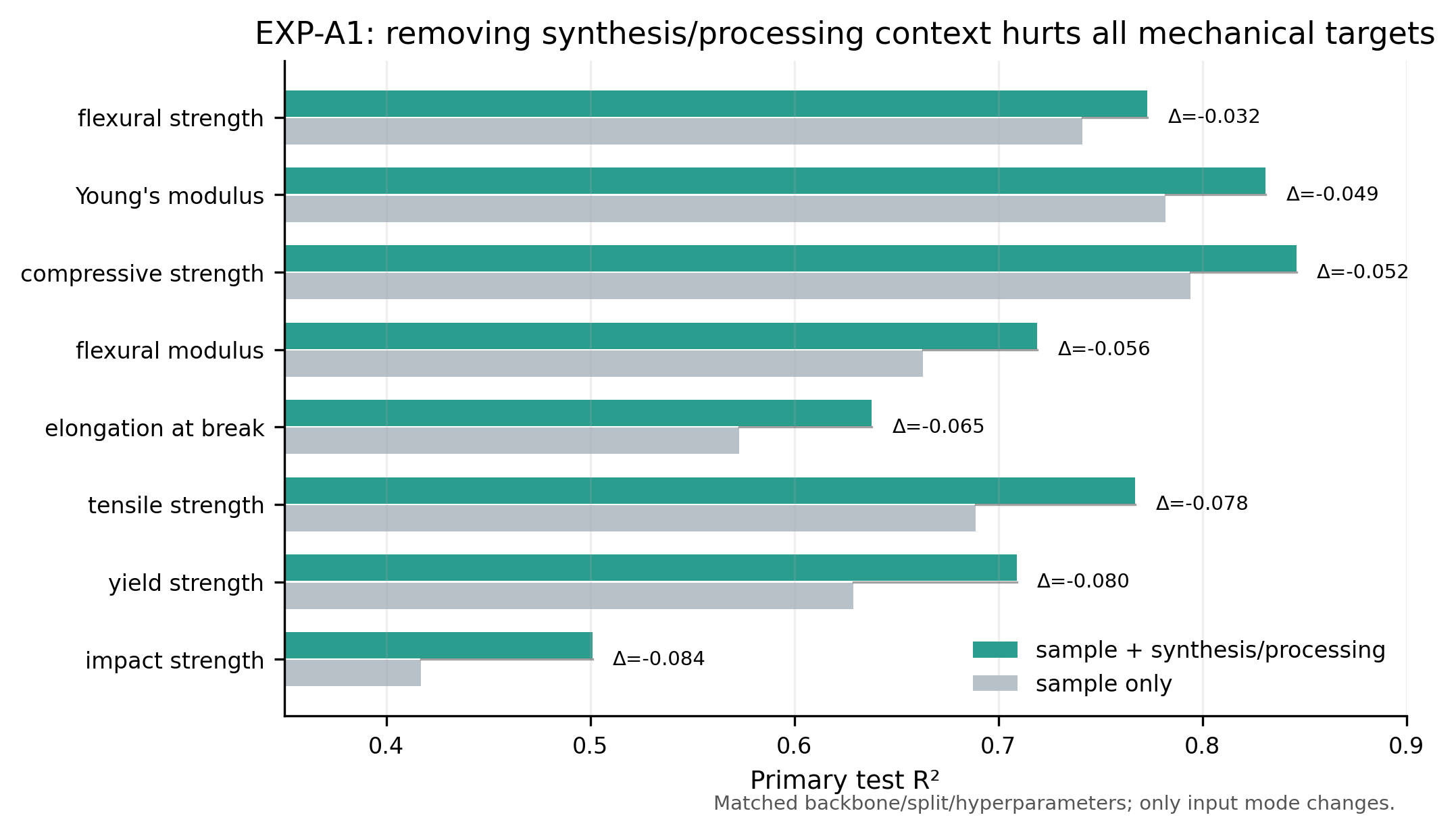}
        \caption{EXP-A1 Mechanical Input Ablation. The chart compares the predictive performance ($R^2$) of the model when provided with full \texttt{sample\_synthesis} context versus a stripped \texttt{sample\_only} baseline. Removing processing context universally harms the prediction of mechanical properties.}
        \label{fig:ablation}
    \end{minipage}\hfill
    \begin{minipage}[t]{0.48\textwidth}
        \centering
        \includegraphics[width=\linewidth]{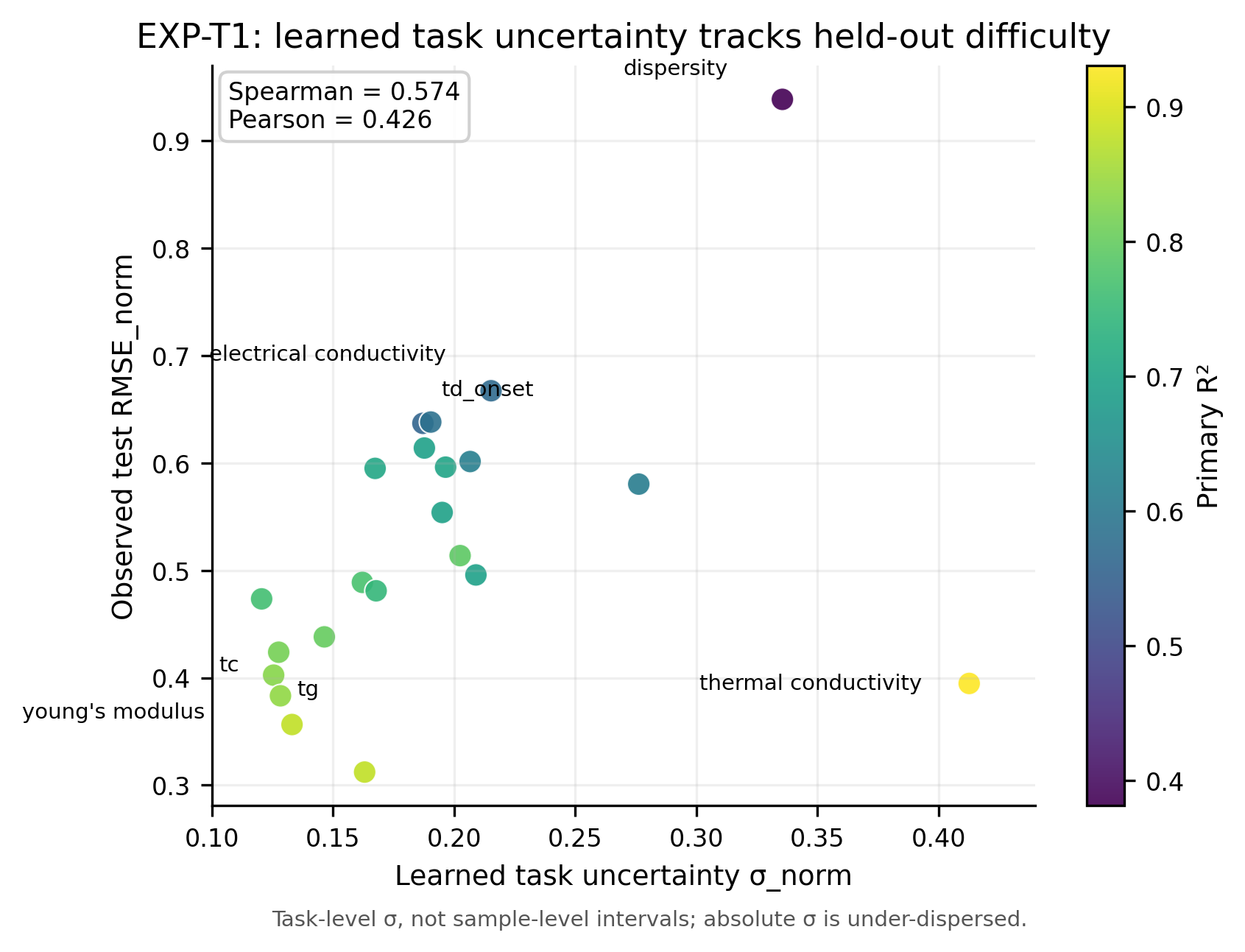}
        \caption{EXP-T1 Task-Level Uncertainty Scatter. The scatter plot illustrates the positive correlation between the normalized learned task uncertainty ($\sigma$) and the normalized empirical test error (RMSE), indicating the model partially captures the inherent difficulty of different property prediction tasks.}
        \label{fig:uncertainty}
    \end{minipage}
\end{figure}

In EXP-T1, we analyze the learned per-property homoscedastic uncertainty parameters derived from the multitask learning objective. Figure~\ref{fig:uncertainty} visualizes this relationship. At the task level, the normalized learned uncertainty correlates positively with the normalized held-out RMSE, yielding a Pearson correlation of 0.426 and a Spearman rank correlation of 0.574. 

The model successfully assigns higher uncertainty weights to inherently noisy or complex targets (e.g., elongation at break, electrical conductivity, dielectric constant) and lower weights to more determinable thermal properties (e.g., $T_m$, $T_g$). However, absolute uncertainties remain significantly under-dispersed (mean RMSE/$\sigma = 2.870$). Thus, while the learned variance parameters function effectively as a task-level difficulty heuristic, they are not calibrated prediction intervals for sample-level inference.

\section{Discussion and Limitations}

Our findings demonstrate that natural language provides a superior modeling interface for unstructured literature data, as mapping records to purely structural SMILES discards vital processing context. Our ablation confirms that this synthesis history is critical for accurate mechanical predictions. Furthermore, the complete zero-shot failure of Claude 3 Opus 4.6 proves that generalist models cannot substitute for our specialized, condition-aware regression architecture.

\paragraph{Limitations}
Despite these advances, natural-language-only modeling is not a panacea. In domains where precise, highly curated molecular graphs are available (e.g., rigid molecular dynamics simulations), structure-based GNNs remain extraordinarily competitive. PolyLM is designed to be complementary, specifically targeting the vast expanse of literature where structural representations are ambiguous or inherently insufficient.

\section{Conclusion}

PolyLM establishes a natural-language-only framework that predicts polymer properties directly from unstructured scientific text. By preserving the crucial synthesis, processing, and testing metadata inherently discarded by structure-only representations, PolyLM achieves high predictive accuracy across 22 diverse properties. Our ablations confirm that this experimental context provides vital mathematical signal, while comparisons against frontier models like Claude 3 Opus 4.6 demonstrate that generalist reasoning cannot substitute for specialized regression. Ultimately, PolyLM champions a paradigm shift in materials informatics: modeling the complete experimental history rather than just molecular identity.

\bibliographystyle{plain}
\bibliography{references}

\end{document}